\documentclass[utf8]{frontiersSCNS} 

\usepackage{url,hyperref,lineno,microtype,subcaption}
\usepackage[onehalfspacing]{setspace}

\usepackage{todonotes}

\def\keyFont{\fontsize{8}{11}\helveticabold }
\def\firstAuthorLast{Vagvolgyi {et~al.}} 
\def\Authors{Balazs P. Vagvolgyi\,$^{1}$, Mikhail Khrenov\,$^{2}$, Jonathan Cope\,$^{3}$, Anton Deguet\,$^{1}$, Peter Kazanzides\,$^{1}$, Sajid Manzoor\,$^{3}$, Russell H. Taylor\,$^{1}$ and Axel Krieger\,$^{1,2,*}$}
\def\Address{$^{1}$Laboratory For Computational Sensing and Robotics, Whiting School of Engineering, Johns Hopkins University, Baltimore, MD, USA \\
$^{2}$Department of Mechanical Engineering, A. James Clark School of Engineering, University of Maryland, College Park, MD, USA  \\
$^{3}$Anaesthesia and Critical Care Medicine, Johns Hopkins Hospital, Baltimore, MD, USA }
\def\corrAuthor{Axel Krieger}

\def\corrEmail{axel@jhu.edu}

\begin{document}
\onecolumn
\firstpage{1}

\title[Telerobotic Operation of ICU Ventilators]{Telerobotic Operation of Intensive Care Unit Ventilators} 

\author[\firstAuthorLast]{\Authors} 
\address{} 
\correspondance{} 

\extraAuth{}

\maketitle

\textbf{7859 Words}, \textbf{9 Figures}, \textbf{1 Table}

\begin{abstract}

\section{}
Since the first reports of a novel coronavirus (SARS-CoV-2) in December 2019, over 33 million people have been infected worldwide and approximately 1 million people worldwide have died from the disease caused by this virus, COVID-19. In the US alone, there have been approximately 7 million cases and over 200,000 deaths.
This outbreak has placed an enormous strain on healthcare systems and workers.
Severe cases require hospital care, and 8.5\% of patients require mechanical ventilation in an intensive care unit (ICU).  
One major challenge is the necessity for clinical care personnel to don and doff cumbersome personal protective equipment (PPE) in order to enter an ICU unit to make simple adjustments to ventilator settings.
Although future ventilators and other ICU equipment may be controllable remotely through computer networks, the enormous installed base of existing ventilators do not have this capability.
This paper reports the development of a simple, low cost telerobotic system that permits adjustment of ventilator settings from outside the ICU.  The system consists of a small Cartesian robot capable of operating a ventilator touch screen with camera vision control via a wirelessly connected tablet master device located outside the room. 
Engineering system tests demonstrated that the open-loop mechanical repeatability of the device was 7.5\,mm, and that the average positioning error of the robotic finger under visual servoing control was 5.94\,mm. Successful usability tests in a simulated ICU environment were carried out and are reported.  
In addition to enabling a significant reduction in PPE consumption, the prototype system has been shown in a preliminary evaluation to significantly reduce the total time required for a respiratory therapist to perform typical setting adjustments on a commercial ventilator, including donning and doffing PPE, from 271 seconds to 109 seconds.

\tiny
 \keyFont{ \section{Keywords:} robotics, telerobotics, coronavirus, ICU, ventilator, PPE,  visual servoing, touch screen} 
\end{abstract}

\section{Introduction}

Since the first reports of a novel coronavirus (SARS-CoV-2) in December 2019, over 33 million people have been infected worldwide and approximately 1 million patients across age groups worldwide have died from the disease caused by this virus (COVID-19) according to the \cite{who_2020}. COVID-19 is a respiratory viral disease with transmission via respiratory aerosols and micro-droplets. This places clinicians and nurses at risk of contracting the virus when caring for patients infected with COVID-19. The primary morbidity and mortality of COVID-19 is related to pulmonary involvement, and according to data from the \cite{cdc_2020}, pneumonia was the primary cause of death in 45.2\% of COVID-19 cases between February 1, 2020 and September 26, 2020 in the US. 8.5\% of patients who develop COVID-19 will require ventilation in an intensive care unit (ICU) at some point during their illness according to a recent meta-analysis by \cite{Chang2020}. 

This pandemic has shown that the scarcest resources necessary to fight COVID-19 are personal protective equipment (PPE), ventilators to combat poor oxygenation, and trained clinical staff. The infection risk for staff and the strain on PPE resources is exacerbated by the fact that for an infectious disease such as COVID-19, healthcare workers must don and doff PPE every time they enter an ICU, even if only to perform a simple task such as changing a setting on a ventilator. Although ICU equipment may eventually be controlled remotely through an in-ICU network, this is not currently the case, and the installed equipment base is not amenable to this solution. 

Medical robots can play a key role in reducing the infectious risk for staff by reducing the amount of close encounters with patients. A recent paper by \cite{Yangeabb5589} categorizes the role of robotics in combating infectious diseases like COVID-19 in four areas, including clinical care, logistics, reconnaissance, and continuity of work/maintenance of socioeconomic functions.
Since the beginning of the pandemic, companies and researchers proposed several such robotic systems for automated temperature screening, \cite{Gong2020}, remote cardiopulmonary imaging, \cite{Ye2020}, taking nasal swabs, \cite{Li2001912} and \cite{swabbot_2020}, autonomous vascular access, \cite{Chen2020}, facilitating rapid COVID-19 testing, \cite{nature_biotechnology_2020}, addressing mental health challenges and supplementing distanced education, \cite{Scassellatieabc9014}, promoting social well-being, \cite{Henkel2020}, and for general telepresence with bimanual teleoperation in ICUs, \cite{Yang2020}.
However, none of the current robotic systems are capable of converting the existing installed base of ventilators and other ICU equipment to remote operation.

\begin{figure}[h!]
\begin{center}
\includegraphics[width=13cm]{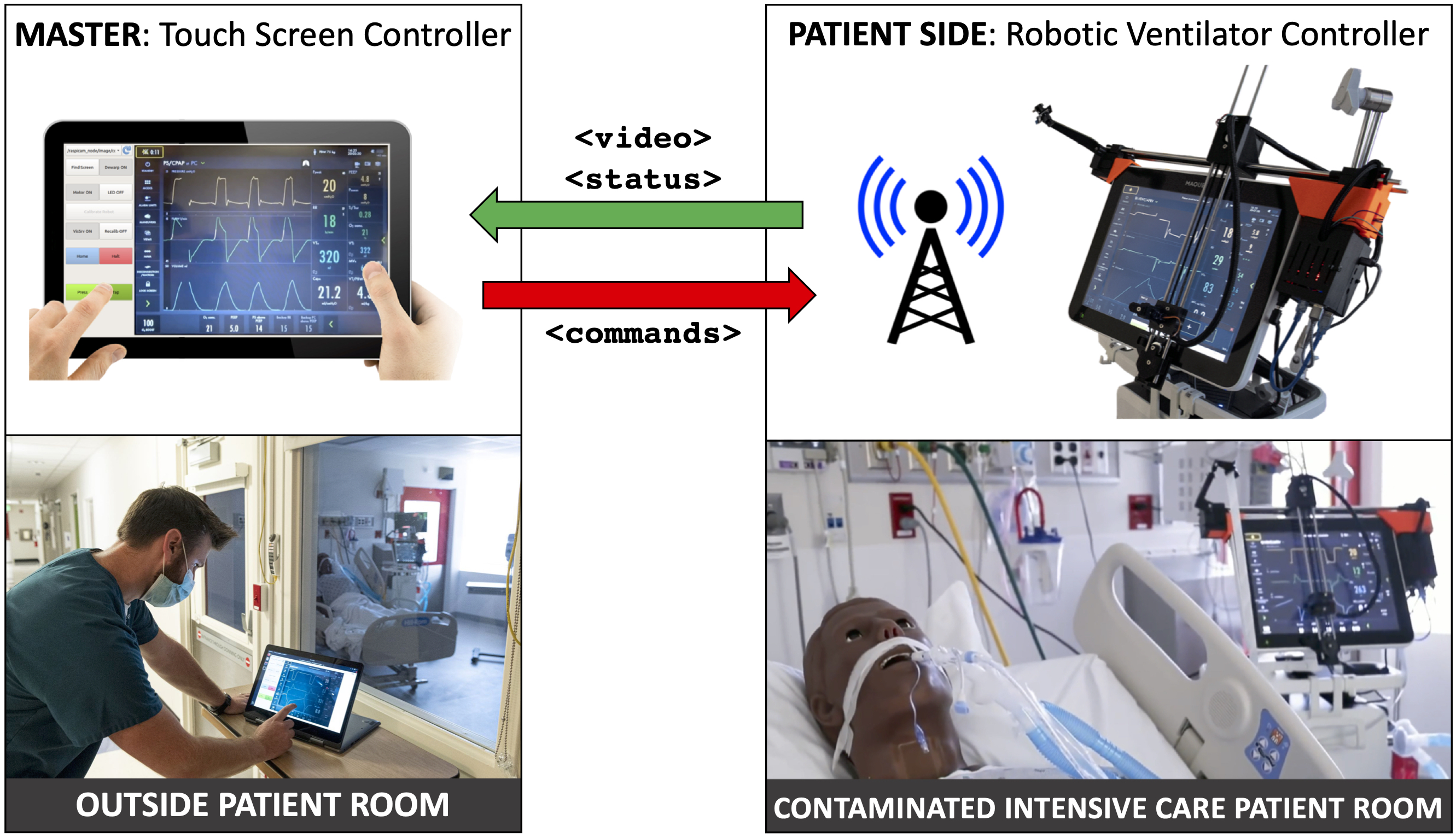}
\end{center}
\caption{System overview: ventilator-mounted camera streams video to master touchscreen; healthcare worker outside patient room presses buttons on video displayed on master touchscreen, which sends commands to ventilator-mounted robot to move to and press buttons on the actual ventilator touchscreen.}
\label{fig:high_level_fig}
\end{figure}

The goal of this work is to develop a rapidly deployable solution that will allow healthcare workers to remotely operate and monitor equipment from outside the ICU room. As shown in Fig. \ref{fig:high_level_fig}, these robots are controlled from outside the ICU by a healthcare worker via a tablet, using encrypted communications to ensure security and patient privacy. Tablet computers are ideal for the healthcare settings because they can be easily cleaned with well-defined infection-control procedures, according to \cite{Hollander2020}.  To meet the urgency of the crisis we prioritized the development and deployment of a remote controlled Cartesian robot dedicated to the most prevalent touchscreen controlled ventilator at Johns Hopkins Hospital (JHH), the Maquet Servo-U, with plans to then expand capabilities and robots to other ventilators and infusion pumps.

Robotic control of touch screens is not unprecedented, but the application of these existing systems is exclusively for touch screen reliability testing in an industrial setting. Such systems include MATT by mattrobot.ai (Bucharest, Romania), SR-SCARA-Pro by Sastra Robotics (Kochi, Kerala, India), and Tapster by Tapster Robotics (Oak Park, IL, USA). All of these systems use a capacitive stylus to interact with the touch screen, but the robot kinematic structures are different from the Cartesian design of our system, utilizing either a delta or SCARA configuration, and completely enveloping the screen they are intended to manipulate.

The primary contributions of this work consist of a custom Cartesian robot designed to interact with a touch sensitive display and a computer vision-based teleoperation method that together effectively enable the replication of the direct interaction scheme with a touch screen on a master tablet console. Further contributions include thorough evaluation of this robotic system with a series of engineering system tests determining open-loop repeatability, closed-loop visual servoing accuracy, and test deployment in an ICU environment. 
As described in Section \ref{discussion}, additional actuator modules that enable the interaction with other physical controls, such as buttons and knobs, have the potential to broaden the range of replicable control interfaces. Applications of such systems range from the safe teleoperation of medical devices in infectious environments to remote management of industrial assembly lines.

\section{Materials and Methods} \label{methods}

The teleoperated ventilator controller system consists of a custom robotic \textit{patient side} device and a touch based \textit{master} console. Computer vision tasks that enable the intuitive user interface and accurate robot control are executed on the \textit{master}. Communication between \textit{master} and \textit{patient side} is implemented in a component-based architecture using the Robot Operating System (ROS), as described by \cite{Quigley2009}.

\subsection{Ventilator-Mounted Cartesian Robot} \label{methods_robot}
The main component of the robot teleoperation system is the robot itself. While  Cartesian robots are nothing new, those available on the market are not optimized for the ventilator touch screen control application. Existing robots are primarily designed for manufacturing or plotting tasks, both of which are performed on steady, horizontal surfaces. As ventilator screens are vertical and liable to be moved in operation, a significantly different design is necessary. Said design must be suitable to the unique mounting situation and optimized for weight, cost, and ease of handling by ICU staff, while still providing suitable accuracy and precision for ventilator manipulation. We have successfully designed and built a lightweight Cartesian robot that attaches to a ventilator screen and enables button pushing through a mechanized robotic finger.

This design consists of a two-axis gantry and a mechanized end-effector finger, with the ends of the horizontal (X) axis used to secure the robot to the desired screen. The vertical (Y) axis assembly is cantilevered on the X axis and translates along it. A roller at the bottom of the Y axis engages the screen bezel and prevents unexpected touch interactions. The two axes are driven by a pair of 45\,Ncm NEMA 17 stepper motors via GT2 timing belts. The use of timing belts and stepper motors for the axes allows the robot to translate quickly to the requested positions and to be easily back-driven by an operator in the case of emergencies. 

The end-effector finger is spring-loaded and controlled by a compact servo turning an eccentric retaining cam. As such, to perform a tap the finger follows a sinusoidal linear motion, with the cam rotating \(110^\circ\) at  \(0.5^\circ/ms\), stopping after the end-effector tip touches the screen, dwelling for 20\,ms, and returning at the same rate. If commanded to press and hold, the finger will perform the first half of the motion and maintain the downwards position until commanded to release, at which point it will retract.

A an inexpensive wide-angle camera observes the ventilator screen and Cartesian robot from an adjustable mount attached to the far side of the X axis. This is used to provide immediate feedback to the operator and robot control system on the status of the ventilator and the position of the end-effector.

\begin{figure}[h!]
\begin{center}
\includegraphics[width=\textwidth]{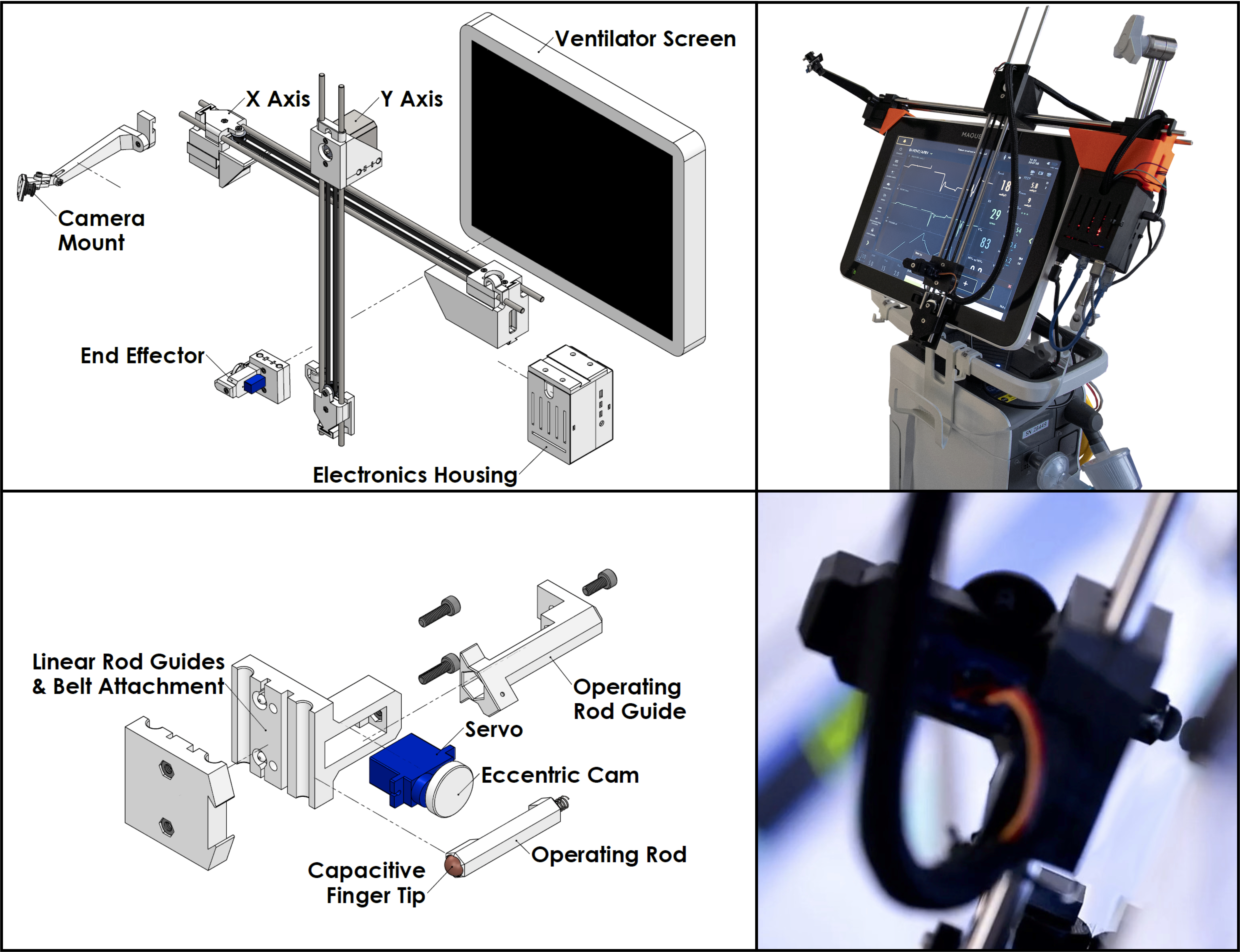}
\end{center}
\caption{The patient side robotic ventilator controller: exploded view drawing of the robotic system (top left), photo of the robotic system installed on a Maquet Servo-U ventilator (top right), detailed exploded view drawing of the end effector assembly (bottom left), close-up photo of the end effector operating a Servo-U ventilator (bottom right).}
\label{fig:robot-cad}
\end{figure}

\begin{figure}[h!]
\begin{center}
\includegraphics[width=0.4\textwidth]{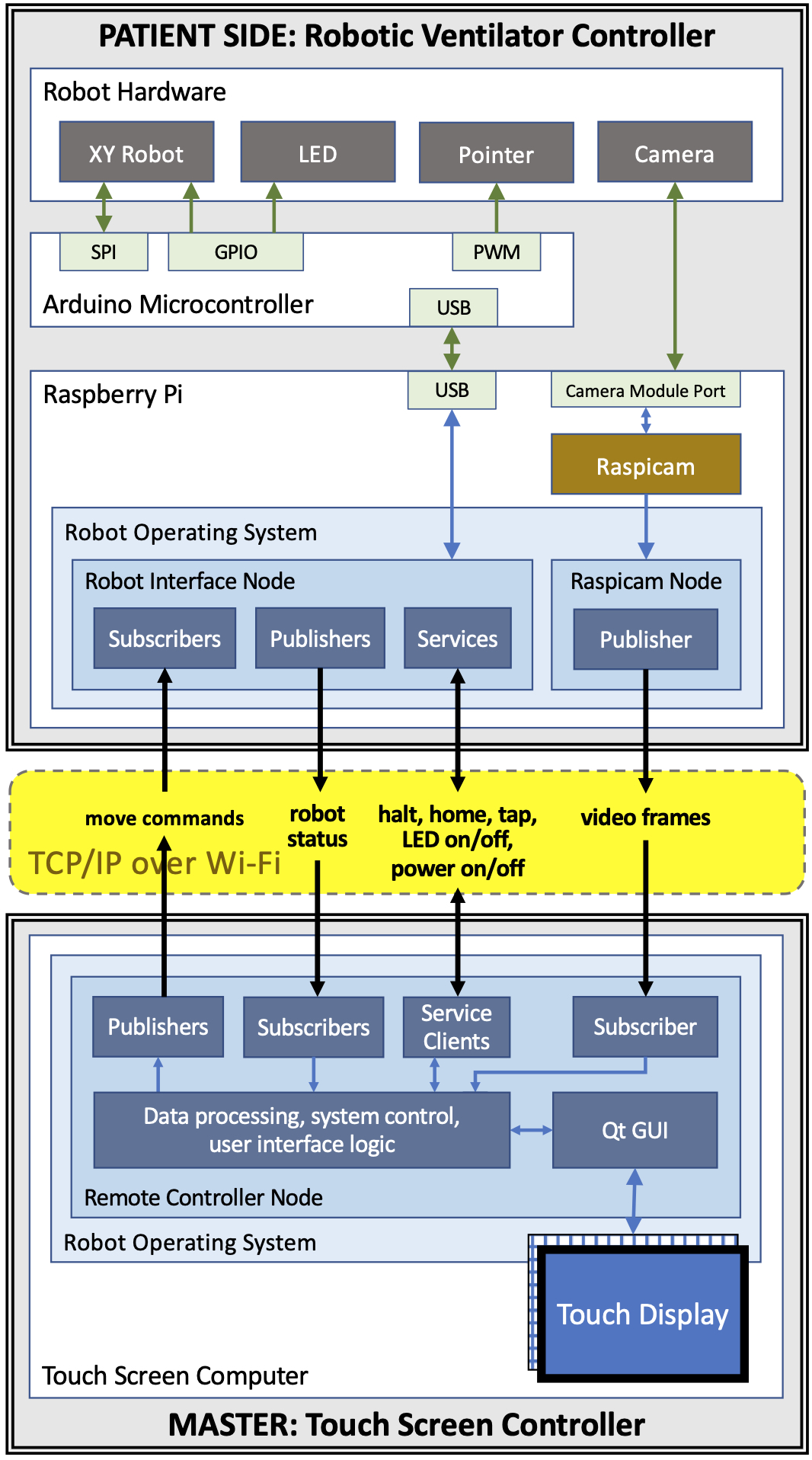}
\end{center}
\caption{Communication channels between components of the teleoperated system: the master and patient side devices communicate with each other using ROS topics and services over WiFi; the patient side Raspberry Pi is connected to the Arduino that controls the motors, the end-effector, and the LED using USB; the camera is connected to Raspberry Pi using the Pi's camera module port.}\label{fig:comm_fig}
\end{figure}

Control for the motors and servo is supplied by an ATmega328 microcontroller alongside TMC2130 stepper motor drivers. The local device firmware, written in C++, takes advantage of the TMC2130 drivers' current sensing capability to perform automatic homing without the use of limit switches and to detect possible collisions with an operator or foreign objects. The microcontroller is connected over a serial port (UART-USB) to a Raspberry Pi microcomputer which provides all the local computing needed in a very light and compact package. Fig. \ref{fig:robot-cad} shows an isometric exploded-view engineering drawing of the design and the assembled robot mounted on a Maquet Servo-U ventilator, while Fig. \ref{fig:comm_fig} illustrates the various communication channels between all hardware components. The Raspberry Pi is connected to the aforementioned camera and provides the network connection needed for the remote controller to drive the robot and monitor the ventilator screen. Parts were bought stock or manufactured via consumer-grade FFF 3D-printing, minimizing weight, cost, and complexity. 

\subsection{Intuitive Robot Control and Visual Servoing} \label{methods_control}

Operators of the proposed system are medical professionals, nurses, respiratory therapists, and physicians with little or no experience with teleoperated robotic systems. Our goal is to make the system easy to operate with very little training by providing a remote-control device with a familiar and intuitive user interface for both setup and operation. We therefore propose a graphical user interface on the remote controller that replicates -- as much as possible -- the appearance of the ventilator's user interface and the way users typically interact with it.

To achieve this, the remote controller device features a large screen on which the live image of the ventilator's control panel is displayed. The live image is captured by a camera placed adjacent to the ventilator inside the ICU.
The optimal angle for the live camera view of the control panel would be provided by a camera mounted directly in front of the ventilator screen (\textit{front-view}). This is not practical, however, because the camera would obscure manual operation of the device and likely interfere with, or be obstructed by, robot motion. It is therefore necessary to mount a camera on the side of the ventilator and use computer vision methods to create an image that replicates the front-view. For some ventilators, it may be possible to obtain the front-view image via an external video output connector, but the side-mounted camera is still required for visual servoing.

For a human user directly operating the ventilator, the brain manages the coordination of hand motions with respect to the visual field.
Conversely, in a robotic remote control system this hand-eye coordination is handled by the robot control algorithm that requires vision feedback to ensure that the robot moves to the correct location and a calibration of the spatial relationships between the camera, the robot, and the ventilator screen.

Our development mainly focused on ventilator models that are controlled exclusively through a touch screen interface, such as the Maquet Servo-U (Getinge AB, Gothenburg, Sweden), but in the Discussion (Section \ref{discussion}) we describe how the system can be modified to accommodate other physical controls, such as buttons and knobs.

\begin{figure}[h!]
\begin{center}
\includegraphics[width=17cm]{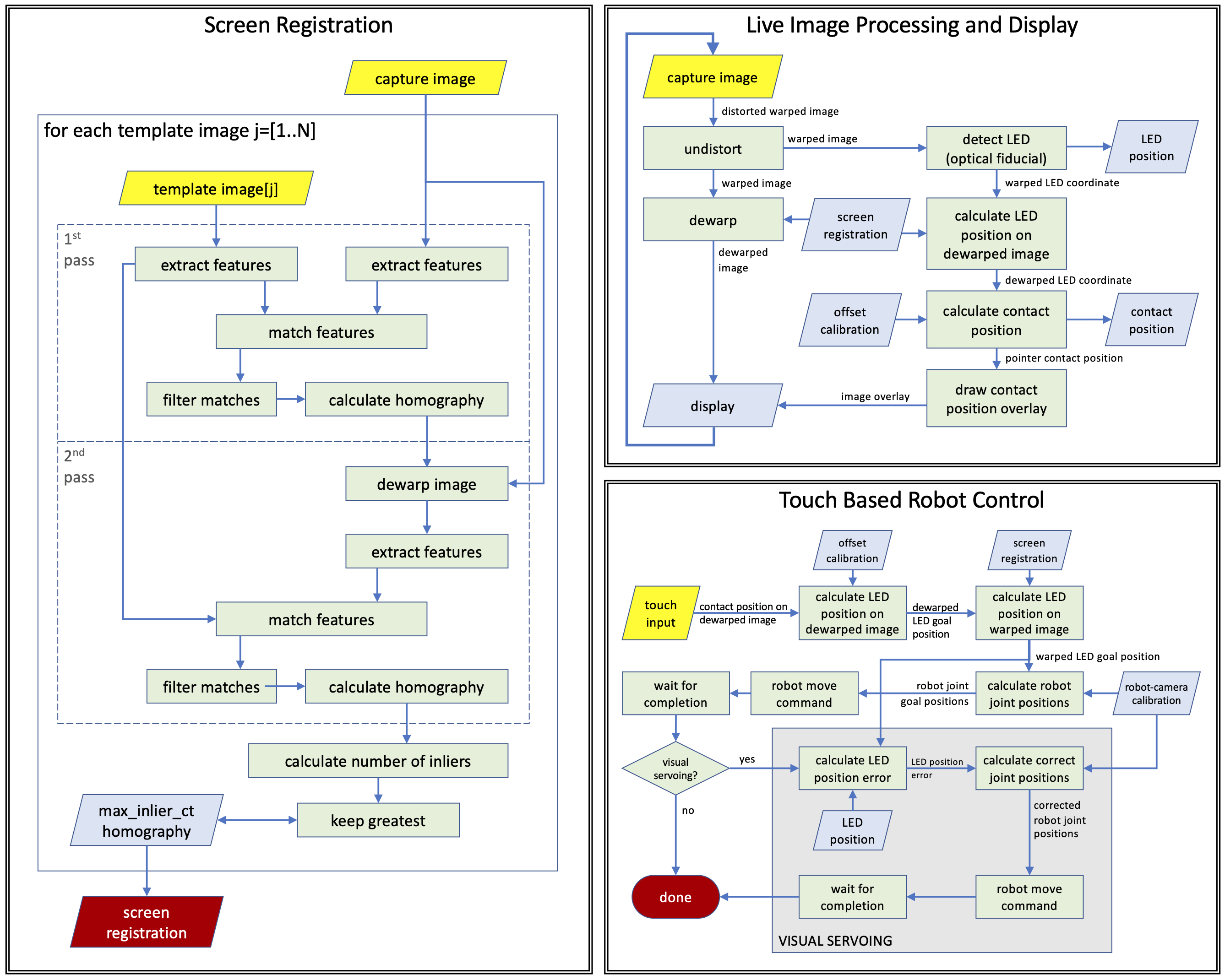}
\end{center}
\caption{Computer vision tasks performed during ventilator screen registration (left), live teleoperation (top right), touch based robot control logic with and without visual servoing (bottom right).}\label{fig:cv_based_control}
\end{figure}

In the following, we describe the components of the vision-based robot control system, all of which, except camera capture, are executed on the remote controller device. Fig. \ref{fig:cv_based_control} illustrates the screen registration method that enables the generation of the front-view, the processing steps performed before displaying a camera frame on the remote controller's screen, and the robot control system's actions in response to a touch event.

\textbf{Image capture:} The camera is mounted on the robot's frame near the upper-left corner of the ventilator screen. Its mounting bracket holds it at 12 cm distance from the image plane, as shown in Fig. \ref{fig:robot-cad}. The current prototype hardware uses an 8-megapixel Raspberry Pi camera module that is configured to capture color images at 10 frames per second in 1920x1440 image resolution. The images are compressed in JPEG format by the ROS \textit{image\_transport} node and sent to the remote controller over a ROS image topic. This particular image resolution was chosen because it provides a reasonable trade-off between spatial fine-detail fidelity, frame-rate, compression time, and bandwidth required. Image quality and framerate were evaluated by a clinical respiratory therapist and were found to be adequate. The intrinsic parameters of the camera were calibrated off-line, which enables the elimination of radial distortion in the first step of vision processing.

\textbf{Visual tracking of robot end-effector:} Live camera images are used by the robot control system to track the robot's position. Knowing the location of the robot's end-effector on video frames enables robot-to-camera calibration and high accuracy robot control by visual servoing.
The visual tracking algorithm is designed to localize a single white light emitting diode (LED) on a dark background and was optimized for real-time performance on a tablet computer when processing a 2.8 megapixel resolution input video stream acquired from the camera. The tracking algorithm first performs image thresholding and connected component analysis to identify large dark areas on the video frames, then uses template matching to find LED candidates in these dark regions. We only search for the LED in dark image regions because the end-effector, on which the LED is mounted, is a black piece of plastic. The template is a 2D Gaussian function that matches the typical size and appearance of the LED on the images. Of the LED candidates, the one with the highest peak intensity relative to its surrounding is selected. There are three special cases that are considered: (1) If the robot is already calibrated to the camera, as described in the next paragraph, then the system assumes that the robot-to-camera calibration is reasonable, which can be used to predict the position of the LED on the image from robot kinematics, and it only looks for the LED on the image in a small neighborhood of the predicted position. This results in a faster and more robust detection. (2) If the two best LED candidates on the image are in a spatial configuration where one of the candidates could be interpreted as a reflection of the LED on the ventilator's screen, then the algorithm will select the candidate that is not the reflection, even if that candidate has a lower peak intensity. (3) If the LED's position is predicted from kinematics to lie outside the visible area, then visual tracking is disabled.

\textbf{Robot-to-camera calibration:} For accurate teleoperation of the system using the remote live view, the robot must be calibrated to the camera, which is done as part of the auto-calibration process of the system. The calibration method moves the robot’s end-effector to 4 or more locations with known joint positions, while the system uses a computer vision method to track the optical fiducial mounted on the end-effector in the camera frames. As the Cartesian frame of the robot is aligned with the ventilator screen, the fiducial will also move along a plane near and parallel to the screen.
During the calibration process, the system stores the end-effector joint positions with the corresponding image coordinates and calculates a homography between the robot and the image coordinates. A homography is suitable for modeling this transformation because the robot's XY joints are prismatic and their scales are linear. The resulting homography enables the mapping between robot joint positions and image coordinates of the optical fiducial in the camera image. The robot calibration does not require operator interaction, takes less than 30 seconds to complete, and is valid as long as the relative position of the camera and the robot is unchanged. In case the relative position of camera and robot should change, the robot calibration process can be executed again remotely without entering the ICU.

\textbf{Fiducial offset calibration:} The location where the LED is mounted on the end-effector was selected to provide good visibility in any allowed end-effector position. Since its position is fixed on the end-effector, it moves rigidly with the capacitive touch device (mechanical finger), but knowing the LED's position in image coordinates is not sufficient to determine where the pointer will touch the screen. In order to be able to calculate that for any end-effector position, a fiducial offset calibration is performed offline, before mounting the robot on the ventilator. Luckily, all input values are coordinates on planes observed under perspective projection, therefore this calibration can also be modeled by a homography. The offset calibration is carried out using a different capacitive touch sensitive display of the same resolution and dimensions as the Servo-U screen, further discussed in Section \ref{results}. This display -- just like the ventilator screen -- has a completely flat glass surface. The identical setup enables us to use the same offset transformation calculated with the calibration display on the ventilator screen. During calibration, we send the robot to predetermined positions in joint space, while tracking the optical fiducial, then we command the robot to touch the screen and record the detected touch coordinates and the corresponding fiducial image coordinates for each position. Finally we calculate the calibration: the homography that describes the transformation between the two sets of coordinates and can also be applied to convert between other touch and fiducial coordinates.

\textbf{Image dewarping to generate front-view image:} The placement of the camera provides an oblique view of the screen that needs to be \textit{dewarped} to a rectangular view before displaying it on the remote-control device’s screen. The dewarping can be modeled as a perspective transform, which is described by a homography. The homography is calculated during the auto-calibration process of the system, by registering the camera image showing the ventilator screen to reference images of the ventilator screen. Since the screen of the ventilator is dynamically changing (it displays plots, numbers, icons, etc.), the reference images are generated from ventilator screen shots by manually masking out non-static regions. As shown in Fig. \ref{fig:cv_based_control}, the screen registration process -- that is repeated for every reference image -- is a two pass method that carries out the following processing steps in each pass: (1) it extracts ORB image features, as described in \cite{orb_features}, on both the reference image and the camera image, (2) calculates the matches between these feature sets, (3) uses RANSAC, by \cite{ransac}, to find the homography that best describes the matches. The steps of processing are identical in the two passes but the parameters for the matching algorithm are different so that the first pass performs a quick coarse alignment, while the second pass performs fine-tuning on the results. In the case of multiple reference images, the match with the highest number of inliers is selected as the best match. The resulting homography is used to convert image coordinates between warped (camera) and dewarped (front) views, and to pre-calculate a dewarping look-up table (LUT) that enables efficient dewarping of every camera frame before displaying on the remote controller's screen.

\textbf{Visual servoing:} The proposed robot control system is designed for robustness by incorporating visual servoing, as illustrated in Fig. \ref{fig:cv_based_control}. During operation, the vision system continuously tracks the position of the end-effector in the live camera video and measures the difference between the robot’s tracked position and the expected position calculated from calibration and robot kinematics. Every time the robot reaches the goal position after a move command, the system compares the visually tracked end-effector position to the goal position and calculates the amount of correction necessary for accurate positioning. If the error is larger than a given threshold, a move command is sent to the robot to execute the correction. The system also integrates the correction in the robot-to-camera calibration to provide better estimates for subsequent moves.

\subsection{User Interface for Teleoperation} \label{methods_gui}

The remote controller of the robotic system is a software designed to be run on a tablet-style computer equipped with a touch screen. In our prototype, we use a Dell Inspiron 14 5000 2-in-1 Laptop that features a 14" touch screen and a keyboard that can be folded behind the screen for tablet-style use. The computer communicates with the robot hardware wirelessly, enabling a completely untethered operation.
During teleoperation, the software's graphical user interface (GUI) fills the remote controller's screen, as shown in Fig. \ref{fig:bcu_photos}, with the camera's live image occupying the entire right side, and Graphical User Interface (GUI) elements located on the left side. As the aspect ratio of the prototype tablet's screen is 16:9 while that of the camera image is 4:3, when the image is scaled to fill the entire height of the screen, there is still room left on the side for the GUI elements without occluding the image.

The right side of the user interface, where the front-view live image of the camera is displayed, shows the  screen of the ventilator. To move the robotic pointer to a particular position on the ventilator screen, the operator taps the same location on the live image of the remote controller's screen. The chain of events triggered in response to this event is illustrated in Fig. \ref{fig:cv_based_control}: first, the robot control algorithm calculates the robot joint positions that correspond to the requested location on the ventilator screen using the transformations calculated during screen registration, fiducial offset calibration, and robot-to-camera calibration. Then, a robot move command is issued to the specified joint positions. Once the robot arrives to the goal position, the visual servoing method calculates the offset between the current robot position and the goal position and issues a corrective move command if needed, that will accurately place the end-effector at the desired position. 

In the prototype remote controller software, the user interface elements on the left side can be divided into two groups: a \textit{setup group} and a \textit{control group}. The \textit{setup group} includes buttons to initiate robot-to-camera calibration, register the ventilator's screen, and switches to enable/disable dewarping, turn the LED on/off, enable/disable visual servoing, and turn motors on/off. The \textit{setup group}'s GUI elements are primarily used for debugging and as such, will be moved into a separate configuration panel or will be hidden from users in the production version of the device.
The \textit{control group} contains the buttons that are the most relevant for operators. The most frequently used button is the \textbf{Tap} button, which sends a command to the robot to perform a single tap action, as described in \ref{methods_robot}. The \textbf{Press/Release} button has two states and enables touch-and-hold actions by dividing the forward and backward motions of the pointer into two separate commands. The \textit{control group} also contains the \textbf{Halt} button to cancel the current motion of the robot and the \textbf{Home} button to move the robot to a side position where it does not interfere with direct operator access to the ventilator screen.

\subsection{Software and Communication} \label{methods_communication}

A teleoperated system relies on the communication method between the master and the patient side device. In a healthcare setting, the success of the entire concept relies on the communication channels being safe, reliable and secure. In the proposed system, the connection also needs to be wireless, as routing a cable out from within an isolated room may not be feasible. This wireless connection must not interfere with existing wireless hospital systems, and since a hospital may want to install multiple instances of the remote controller in a unit, the additional wireless links should not interfere with each other either.

In our design, we chose to use industry standard WiFi (IEEE 802.11g-2003 or IEEE 802.11n-2009) connections with built-in WPA2 authentication (IEEE 802.11i-2004). Each instance of the remote controlled system uses its own dedicated WiFi network with only the master and the patient side device being part of the network. This communication method is safe, as it is compatible with hospitals' own wireless systems, secure and highly reliable. It is also easy to deploy and WiFi support is already built into most modern computers.
Having the UDP and TCP protocols available on WiFi networks enabled us to use the Robot Operating System (ROS) for establishing data connections between the master and patient side device. The ROS middleware provides a communication software library and convenient software tools for robotics and visualization. ROS communication is not secure in itself but channeling its network traffic through a secure WiFi network makes our system secure.
The communication channels between the components of the teleoperated system are shown in Fig. \ref{fig:comm_fig}.

Both the master and the patient side of our system run on the Linux operating system. The patient side Raspberry Pi runs Raspbian Buster while the master runs Ubuntu 18.04. Both systems have ROS Melodic installed. The GUI software on the master is implemented in C++ using RQT, a Qt based GUI software library with access to the ROS middleware. The patient side software is also implemented in C++. Computer vision methods were implemented using the OpenCV software library, \cite{opencv_library}.

\section{Results} \label{results}
To quantify the effectiveness of the robotic teleoperation system, four experiments were carried out, intended to measure the open-loop repeatability, the closed-loop visual servoing accuracy, relative time needed to operate a ventilator with the system, and the qualitative user experience of the teleoperation system.

As, due to the COVID-19 pandemic, access to mechanical ventilators for testing was precious and limited, a ``mock ventilator'' was constructed to perform the tests, which did not necessitate exact replication of a clinical environment. This mock ventilator consisted of a commercially available point-of-sale capacitive touch screen monitor, selected to match the Servo-U screen size and resolution, connected to a laptop computer running test software written in JavaScript. The software was programmed to display screen captures from the Servo-U ventilator, record the location of any touch interactions in screen pixel coordinates, and emulate three commonly used features of the Servo-U: changing the oxygen concentration, activating the oxygen boost maneuver, and changing the respiratory rate alarm condition.


\subsection{Mechanical Repeatability Testing}
In order to quantify the open-loop mechanical repeatability of the robotic system, we performed a positional repeatability test according to \cite{ISO9283} (\textit{Manipulating industrial robots — Performance criteria and related test methods}). The robot was commanded to move in sequence to 5 positions in joint space distributed across the screen, performing the sequence a total of 50 times. At each position, the robot paused and tapped the screen, with each touch location on the screen being registered in pixel coordinates by the test software. Given the measured dimensions of the screen and its defined resolution (1024 by 768 pixels), the pixel width and height were both found to be approximately 0.3 mm, which enabled the conversion of pixel coordinates to position coordinates in millimeters.

\begin{figure}[h!]
\begin{center}
\includegraphics[width=\textwidth]{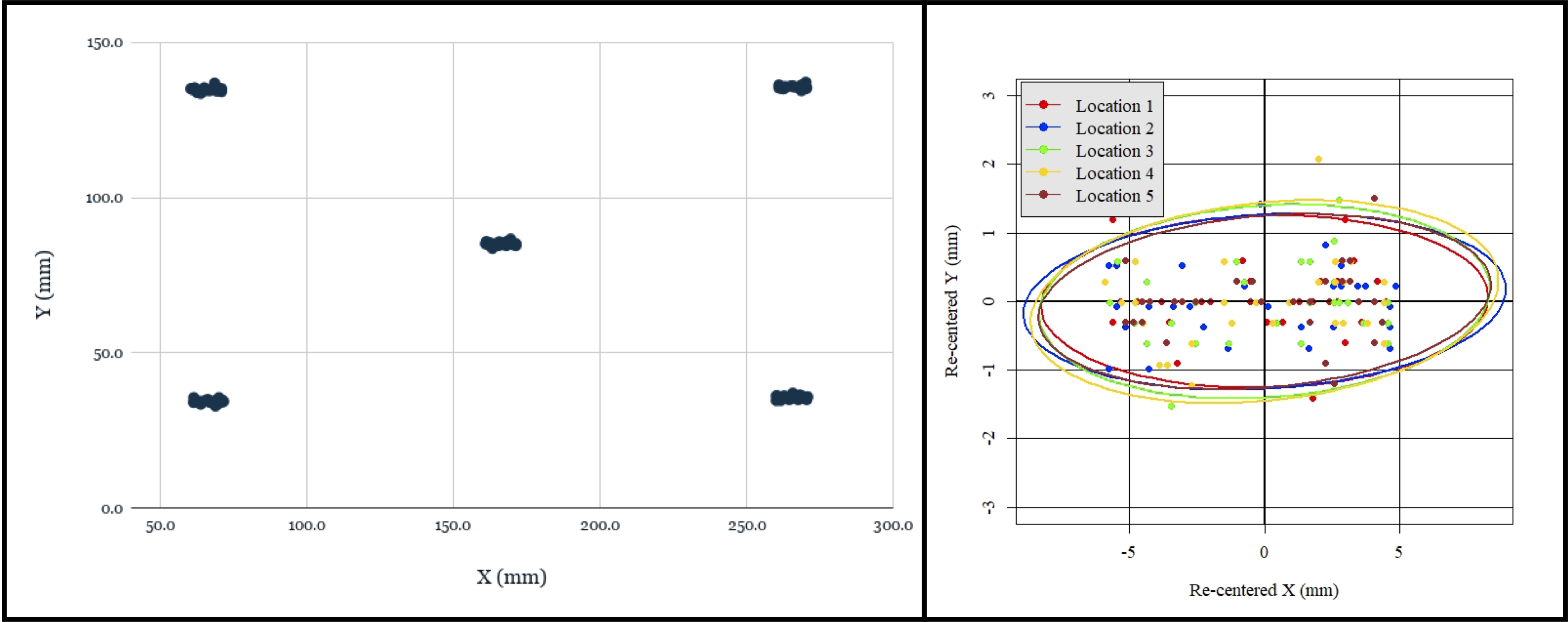}
\end{center}
\caption{Evaluation of robot open loop repeatability: Scatter plot of open-loop recorded taps for 5 locations distributed across the screen (left), scatter plot of all tap locations re-centered with respect to each barycenter with 95\% confidence covariance ellipses for each location. Notably, the spread of re-centered locations is approximately the same across all locations, as evidenced by the covariance ellipse overlap, but spread is greater in X than in Y.}
\label{fig:open-loop-scatter}
\end{figure}

Figure \ref{fig:open-loop-scatter} shows the location data from the 5 groups of 50 taps, re-centered about their respective barycenter. As can be seen, the distributions were uniform and practically indistinguishable across the 5 locations. The pose repeatability (RP), is defined by \cite{ISO9283} as \(RP = \bar{l} + 3S_l\), for \(\bar{l}\) average euclidean distance to barycenter, and \(S_l\) standard deviation of euclidean distances to barycenter. For the 250 position measurements taken from the robotic prototype, RP was found to be 7.5mm. While these results are far from high precision positioning, given that the smallest button on the Servo-U ventilator needed for setting adjustment is 21\,mm by 21\,mm, it is appropriate for the task at hand.

The one notable feature of the data was that the spread was significantly greater in the X (horizontal) direction than the Y (vertical) direction, ranging $\pm$5\,mm in X and only $\pm$1.5\,mm in Y. The reason for this was readily apparent: due to its low-cost and lightweight construction, the prototype design omitted linear bearings, with the two axis frames riding directly on the linear rods with a loose 3D-printed slip-fit cylindrical feature. Due to the play in this feature and the cantilevered design, the Y axis arm could swing a small angle, inducing errors. Due to the arm's length and the small degree of the swept angle, this issue's impact on the Y repeatability was minimal (as it was of order \(cos(\theta) \approx 1\)), but the impact in the X direction was detectable (\(sin(\theta) \approx \theta\)). This issue could be resolved in future prototypes with the use of close-fitting linear bearings.

\subsection{Visual Servoing Accuracy Testing}
\begin{figure}[h!]
\begin{center}
\includegraphics[width=\textwidth]{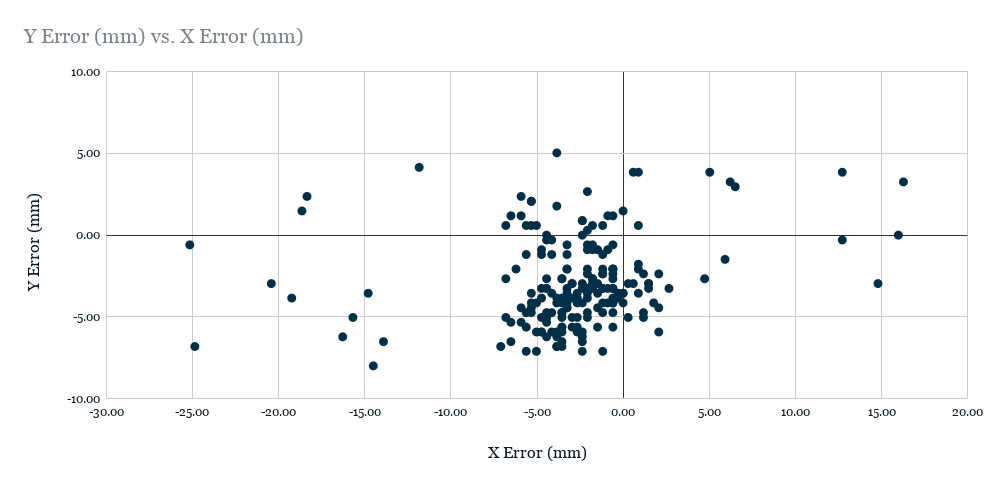}
\end{center}
\caption{Evaluation of robot positioning accuracy: Scatter plot of visual servoing recorded tap location errors for the 40 target locations spanning the screen across the 5 experimental runs (200 total points). See discussion.}
\label{fig:closed-loop-scatter}
\end{figure}

In order to quantify the accuracy of visual servoing, 40 uniformly distributed positions were generated across the ventilator screen. In a random sequence, each of these locations was displayed on the mock ventilator screen by means of a thin black crosshair on a white background. The experimenter, using the teleoperation interface, would then command the robot to move (with visual servoing enabled) to the center of the cross-hair, using a mouse to ensure precise selection. Upon arriving at its destination, the robot would be commanded to tap, with the resulting pixel coordinates being recorded by the test software of the mock ventilator. This sequence was repeated 5 times, each time using a different screenshot taken directly from the Servo-U ventilator for screen registration. The raw errors for all 40 locations across the 5 runs are shown in Figure \ref{fig:closed-loop-scatter}.

The X average error and Y average error were found to be -2.87\,mm and -2.89\,mm, respectively, with standard deviations of 5.31\,mm and 2.71\,mm.
The average Euclidean error was found to be 5.94\,mm with a standard deviation of 4.19\,mm,  where 89.5\% of the data points (179 of 200) are clustered around the barycenter within 2$\sigma$ radius, with the remaining points considered outliers. The analysis of the results suggest that the factors responsible for the errors for the data points within the cluster are related to mechanical precision, system calibration, and screen registration inaccuracies, while the outliers were produced as a result of failed vision-based tracking of the optical fiducial.

\begin{figure}[h!]
\begin{center}
\includegraphics[width=\textwidth]{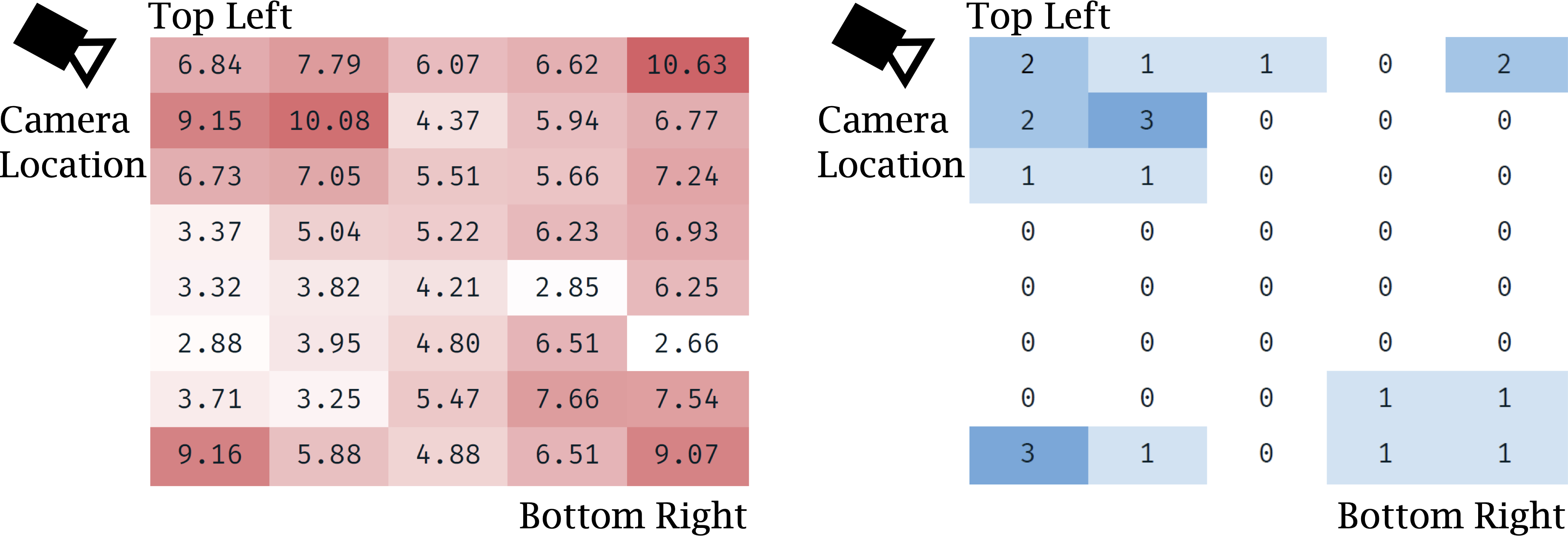}
\end{center}
\caption{Evaluation of robot positioning accuracy with visual servoing: heat map of Euclidean error (in mm) for visual servoing for each of the 40 locations spanning the screen, averaged over the 5 trials (left), heat map of the number of outliers with respect to Euclidean error for each of the same locations spanning the screen, summed over the 5 trials (right).}
\label{fig:closed-loop-heatmap}
\end{figure}

Notably, the error of visual servoing is not uniform across the screen. Figure \ref{fig:closed-loop-heatmap} shows a heatmap of error and the number of outliers across the 40 uniformly distributed spanning locations. Error is overall greatest near the corners and the edges of the screen that are farthest from the view of the left-side mounted camera. Outliers are clustered near the corners, which shows that the vision-based tracker often failed to detect the optical fiducial when the LED was far from the center of the screen. This failure is particularly apparent in the top left corner, nearest the camera, which suggests that in that particular configuration the LED detection algorithm often confused the LED with another nearby bright spot in the camera's image. The larger errors in the other 3 corners, farther from the camera, are partially due to the lower spatial fidelity of the image at those locations, which leads to significantly higher errors when measured as projections on the image plane.

While an average 5.94\,mm of Euclidean positioning error is significant, it is sufficient for the teleoperation tasks required of the prototype given the aforementioned 21\,mm minimum feature size. Further planned improvements for the visual servoing system are discussed later in this text (Section \ref{discussion}).

\subsection{Manual Operation vs. Teleoperation Setting Change Time Comparison}
\label{results_timed}
To verify the usability and utility of the device, the prototype teleoperated Cartesian ventilator robot was mounted on the mock ventilator. Experimenters were asked to perform three tasks representative of routine setting changes, manually and via teleoperation: increasing the oxygen concentration setting by five percentage points, activating the O2 boost maneuver, and lowering the respiratory rate alarm condition by three increments. The experiments were recorded and timed from the first interaction to the confirmation of the last setting change. Table \ref{tab:times} shows the results for three such experiments performed on the mock ventilator, and one experiment performed on an actual Servo-U.

The data showed that, on average, operators were able to complete the three tasks in 18\,s manually and in 67\,s via teleoperation, using a mock ventilator screen. They were able to complete the same tasks in 28\,s manually versus 109\,s using a clinical ventilator, a ratio of approximately 3.8 in terms of additional time needed using the robot. However, performing the tasks manually in a clinical setting would require a respiratory therapist to fully don PPE. A co-investigator with clinical expertise performed the full don/doff sequence that would be required and recorded the times as being 170\,s to don and 73\,s to doff, not including time to clean equipment post-doff. Thus, in total, including donning and doffing, manual operation of the Maquet Servo-U ventilator for these tasks would have taken 271\,s compared to only 109\,s for teleoperation with our prototype Cartesian ventilator robot, leading to a significant net time savings, in addition to reducing the amount of PPE consumed and the risk of infection to the respiratory therapist. Due to limited time and access to hospital resources during pandemic conditions, we have only a single data point for donning and doffing times and for teleoperation times with the clinical ventilator, therefore our results are not yet conclusive. However, our clinical collaborators confirmed that these times are representative of typical ventilator operation and PPE donning and doffing times.

\subsection{Qualitative System Evaluation in Biocontainment Unit} \label{results_bcu_evaluation}

\begin{figure}[h!]
\begin{center}
\includegraphics[width=17cm]{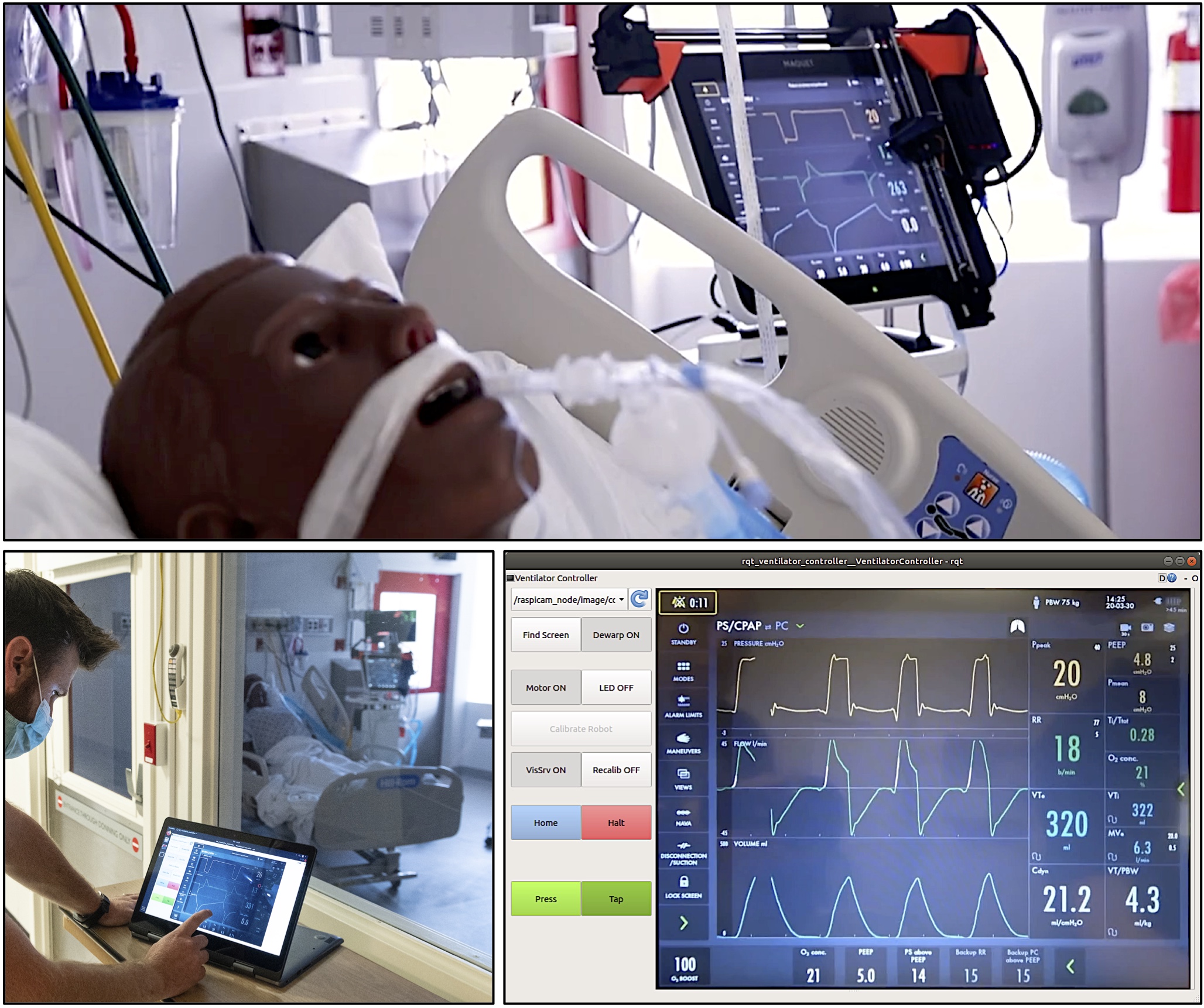}
\end{center}
\caption{Evaluation of the ventilator controller system on a human body phantom in the Johns Hopkins Hospital's biocontainment unit (BCU). Robot mounted on the screen of a Maquet Servo-U ventilator (top), respiratory therapist using the remote controller software on a tablet computer outside the BCU (bottom left), remote controller user interface layout (bottom right). }\label{fig:bcu_photos}
\end{figure}

We took the teleoperated ventilator controller system to the Johns Hopkins Hospital's Biocontainment Unit (BCU) for qualitative evaluation in maximally accurate conditions. The tests were carried out by a clinical respiratory therapist (RT) while the patient side robotic manipulator was mounted on a Maquet Servo-U mechanical ventilator, as shown in  Fig. \ref{fig:bcu_photos}. The ventilator was connected to a human body respiratory phantom to enable the simulation of realistic usage scenarios. Before evaluation, the camera and robot were calibrated using the system's remote controller, after which the RT spent approximately two hours using the system for making typical adjustments on the ventilator from outside of the BCU bay (Fig. \ref{fig:bcu_photos} on bottom left). The wireless signal easily penetrated into the BCU bay resulting in reliable communication between the remote controller and robot. 

During and after the evaluation, the RT provided invaluable feedback regarding the system. His feedback is summarized by topic in the following: \\
\textbf{Image quality:} The quality of the live front-view camera image was assessed to be adequate to read numerical information and plots displayed on the ventilator's screen. The respiratory therapist found no issues with the frame rate of 10 frames per second provided on the remote controller's screen. He suggested that since the most relevant information on the ventilator's display is on the right side, the remote controller would be able to provide a higher definition view of that information if the camera was moved from the top left to the top right corner. He found the current video latency of approximately 1 second distracting, and recommended that it should be reduced in a production version of the system. \\
\textbf{Robot:} The RT emphasized that the robot may not yet be physically robust enough to be used in a healthcare setting and would need to be ruggedized. He also asked us to investigate if the robot being mounted on the ventilator screen can potentially affect the range of motion of the ventilator's swiveling display and whether the robot's mounting points are compatible with other ventilator models with different screen thickness. \\
\textbf{Graphical user interface (GUI) of the remote controller:} The current prototype remote controller's graphical user interface contains buttons and switches that are for debugging purposes, which the RT suggested should be hidden or moved into a dialog box. He also mentioned that the current user interface is inconvenient for right-handed operation and we should add an option for switching between left and right-handed layouts. The most significant feedback regarding the GUI we received is that the location of the \textbf{Tap} button is unintuitive in the current fixed position and it would be preferred to have it move together with the robot's end-effector near the touch point. He suggested that the \textbf{Tap} button should be merged with the \textbf{Press/Release} button. \\
\textbf{Robot control and visual servoing:} According to the RT, controlling the robot by touch on the remote controller's interface is intuitive and visual servoing seemed to make a positive impact, but due to the video latency of approximately 1 second, visual servoing adds an additional delay that should be reduced. \\
\textbf{On handling multiple systems in a single unit:} The patient identifier should be clearly shown on the remote controller. Currently, ventilator screens do not show patient identifying information, but a remote controller unit that can work from a distance must have the patient ID prominently shown in order to make sure the right ventilator settings are entered for the correct patients.

\begin{figure}[h!]
\begin{center}
\includegraphics[width=18cm]{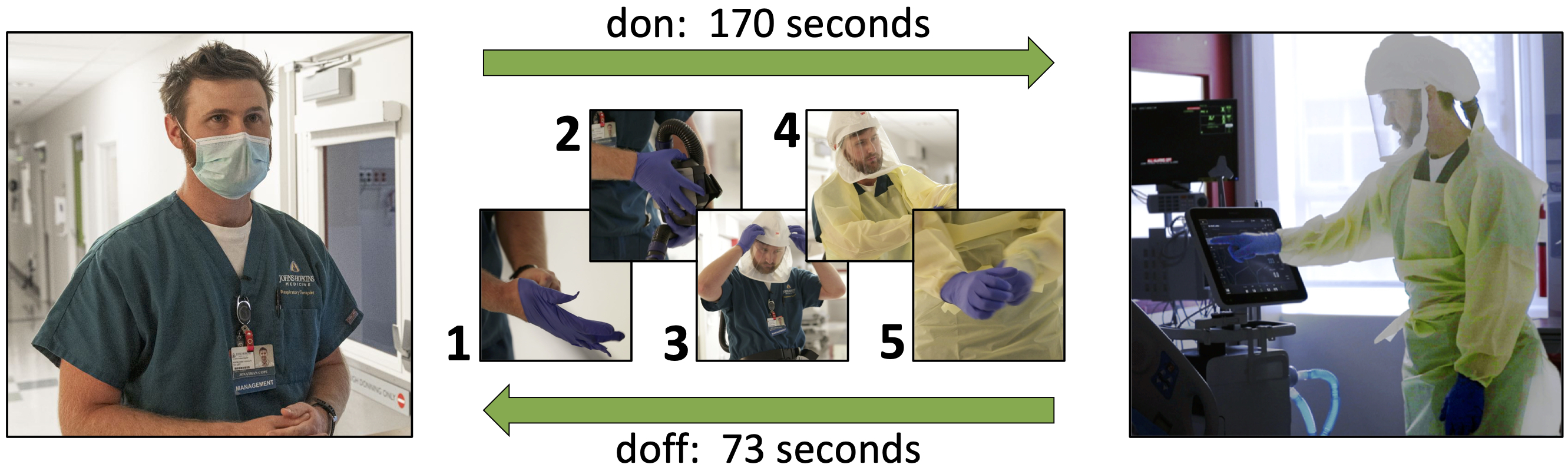}
\end{center}
\caption{During our qualitative system evaluation in the biocontainment unit of the Johns Hopkins Hospital, it took 170 seconds for a respiratory therapist to don his personal protective equipment (1: a pair of disposable nitrile gloves, 2: respirator device, 3: mask, 4: plastic gown, 5: a second pair of disposable nitrile gloves) and 73 seconds for doffing.}\label{fig:don_doff}
\end{figure}

With the help of the RT, as previously mentioned in Section \ref{results_timed} we also measured the time required to don and doff personal protective equipment (PPE) for a healthcare worker to access a negative-pressure intensive care patient room or a BCU from outside. As shown in Fig. \ref{fig:don_doff}, it took 170 seconds for the RT to don and 73 seconds to doff the PPE during the one trial we had a chance to observe. The required PPE included two pairs (two layers) of nitrile disposable gloves, a respirator device with the attached mask, and a plastic gown.

\section{Discussion} \label{discussion}


This paper reports the development of a simple, low cost telerobotic system that allows adjustment of ventilator settings from outside the ICU.  Our experiences with our initial prototype are very encouraging and provide a basis for further development.
Engineering system tests demonstrated that open-loop position repeatability was 7.5\,mm and that the average positioning error of the robotic finger under visual servoing control was 5.94\,mm. Successful usability tests in a simulated ICU environment were also reported. Preliminary evaluation highlighted the system's potential to save significant time and PPE for hospitals and medical staff. In one evaluation where we compared the time required to make an adjustment on the ventilator using the proposed teleoperated system to the time required for a respiratory therapist to don PPE, enter the patient room, make the change directly on the ventilator, then leave the room and doff the PPE, we found that the RT managed to make the adjustment in a significantly shorter time (109\,s) using teleoperation than without (271\,s). We also received positive feedback during qualitative evaluation in a clinical setting. The respiratory therapist who performed the evaluation emphasized that the system can be a force-multiplier for respiratory teams by freeing up valuable resources.
This robotic system has the potential to reduce the infection risk for healthcare workers, reduce usage of PPE, and reduce the total time required to adjust a ventilator setting.

One limitation of the current prototype robotic system hindering clinical usability is the difficulty to adequately clean and disinfect the Cartesian robot. A future clinical grade device will be encased in an acrylic cover to protect recessed features and components from contamination, thus facilitating easier cleaning by wiping down the convex outer surface. Disinfection of the device will follow the CDC’s Environmental Cleaning strategy, WB4224, which requires disinfecting the device with EPA-registered hospital disinfectant during regular cleaning cycles, between patients, and before removing it from the room. 

A second limitation of the reported study was the frequent large position error under visual servoing. 17 of the 200 test taps had an Euclidean error $>$10\,mm, which would result in a miss when tapping small features on the ventilator, which have a minimal size of 21\,mm. While any setting changes on the ventilator require tapping a confirmation button, which prevents accidental parameter adjustments, this high fail rate reduced usability and confidence in the prototype system. There were three contributing factors to the errors that will be addressed in a future clinical system:
(1) The mechanical play in the system created an unintended swing of the end-effector, contributing to the error in the X direction. This error can be easily reduced by using linear bearings.
(2) The visual servoing control did not work robustly due to missed detections of the optical fiducial on the end-effector, which was the root cause of outliers in the visual servoing evaluation results. We found that the detection of a single LED mounted on the end-effector sometimes proved challenging as the LED detection method occasionally detected a different bright spot in the image (e.g., the LED’s reflection on the screen, a similar sized bright dot shown on the ventilator screen, or metallic reflection from a robot component), therefore in the next version of the prototype we will replace the single LED with a small cluster of LEDs arranged in a unique pattern that is distinguishable from its mirror image.
With the use of multiple LEDs we aim to achieve better than 99.5\% success rate in unambiguously determining the end-effector position. While it would be desirable to achieve 100\% detection rate, occasionally missed detections do not invalidate the system as the operator can always manually correct for robot positioning inaccuracies before giving a tap command.
(3) Calibration and screen registration inaccuracies contributed to an error of approximately 4 mm in the visual servoing experiments. This error manifested itself as a consistent offset in touch positions. While this offset alone would not significantly impact system performance, considering the large size of buttons on the ventilator screen, we will reduce the introduced offset by improving the fiducial offset calibration process and screen registration.

During qualitative evaluation of the system in a BCU, our clinical collaborators uncovered usability issues in the graphical user interface of the remote controller software that will be addressed in the next versions of the system. These improvements include the relocation of the interaction buttons, \textbf{Tap} and \textbf{Press/Release}, adding an image zoom feature, and adding an option to allow switching between right-handed and left-handed layouts.

A system used in a clinical setting will also need to properly address patient identification. Currently mechanical ventilators do not require patient identifying information during setup and operation because the device is placed near the patient and it is always entirely unambiguous to which patient the ventilator is connected. However, in a remote controlled scenario, particularly in the situation when there are multiple remote controlled systems used simultaneously with different patients, a lack of patient ID could lead to confusion where adjustments are made to the wrong patients by mistake. We will address this critical issue by adding a patient identifier to the remote controller's graphical user interface and make the entering of patient ID a mandatory step in the system setup process.

To broaden the pool of ventilators that the robot can control, we are planning to incorporate a sub-system to turn an adjustment knob. Both the Hamilton (Hamilton Medical Inc., Reno, NV, USA) ventilator series, which represent the second largest install base at Johns Hopkins Hospital after the Maquet Servo-U, and the GE (General Electric Healthcare Inc., Chicago, IL, USA) ventilators have an integrated knob. The system will use a stepper-driven, spring-loaded friction wheel, which will run along and turn the setting adjustment knob, while the robotic finger will be used to push the confirmation button on the touch screen. 

Our team of clinicians and engineers came together during an extraordinarily challenging time, at the beginning of a global pandemic, to leverage our expertise and experience in assistance of frontline healthcare workers. Team members were geographically separated as a result of social distancing, which made some aspects of system integration and evaluation particularly difficult. Software developers never got to see the hardware in person that they were developing for, and none of the people participating in implementation were able to be present for the system evaluation in the Biocontainment Unit. Despite the difficulties, our initial prototype performed well during evaluation and received mainly positive feedback from clinical professionals. Our team is committed to carry on and make the system sufficiently robust for controlled clinical studies in the near future.




\section*{Conflict of Interest Statement}

The authors declare that the research was conducted in the absence of any commercial or financial relationships that could be construed as a potential conflict of interest.

\section*{Author Contributions}


BPV developed the computer vision and visual servoing algorithms, along with the robot control software and user interface used in this system.
MK performed the detailed design and construction of the prototype robot, wrote the local controller firmware, and conducted the engineering system validation. 
JC provided key insights into system requirements and conducted the usability evaluation.
AD provided key elements of the system software implementation.
PK contributed to the system design.
SM provided expert advice on mechanical ventilation and PPE, as well as key insights into the system requirements.
RHT coordinated the overall effort and contributed to all phases of the design and evaluation discussions.
AK led the development of the in-ICU robot and robot integration, designed the system evaluation studies, and performed the manual vs. teleoperation comparison and BCU studies.
All authors contributed to the article and approved the submitted version.

\section*{Funding}
The reported work was supported by Johns Hopkins University internal funds and University of Maryland internal funds.

\section*{Acknowledgments}

We thank Dr. Brian Garibaldi and the BCU team for access and support for the BCU evaluation; Dr. Sarah Murthi and Dr. Michelle Patch for early discussions on clinical needs and physician perspectives; Anna Goodridge for early design consultation; and Lidia Al-Zogbi and Kevin Aroom for participating in the evaluation of the robotic system. 




\bibliographystyle{frontiersinSCNS_ENG_HUMS} 
\bibliography{frontiers}

\begin{thebibliography}{19}
\providecommand{\natexlab}[1]{#1}
\expandafter\ifx\csname urlstyle\endcsname\relax
  \providecommand{\doi}[1]{doi:\discretionary{}{}{}#1}\else
  \providecommand{\doi}{doi:\discretionary{}{}{}\begingroup
  \urlstyle{rm}\Url}\fi
\providecommand{\selectlanguage}[1]{\relax}
\providecommand{\bibAnnoteFile}[1]{%
  \IfFileExists{#1}{\begin{quotation}\noindent\textsc{Key:} #1\\
  \textsc{Annotation:}\ \input{#1}\end{quotation}}{}}
\providecommand{\bibAnnote}[2]{%
  \begin{quotation}\noindent\textsc{Key:} #1\\
  \textsc{Annotation:}\ #2\end{quotation}}

\bibitem[{{Biobot Surgical Pte Ltd}(2020)}]{swabbot_2020}
[Dataset] {Biobot Surgical Pte Ltd} (2020).
\newblock {SwabBot - The World's First Patient-Controlled Nasopharyngeal Swab
  Robot}
\bibAnnoteFile{swabbot_2020}

\bibitem[{Bradski(2000)}]{opencv_library}
Bradski, G. (2000).
\newblock {The OpenCV Library}.
\newblock \emph{Dr. Dobb's Journal of Software Tools}
\bibAnnoteFile{opencv_library}

\bibitem[{Chang et~al.(2020)Chang, Elhusseiny, Yeh, and Sun}]{Chang2020}
Chang, R., Elhusseiny, K.~M., Yeh, Y.-C., and Sun, W.-z. (2020).
\newblock {COVID-19 ICU and mechanical ventilation patient characteristics and
  outcomes - A systematic review and meta-analysis}.
\newblock \emph{medRxiv} \doi{10.1101/2020.08.16.20035691}
\bibAnnoteFile{Chang2020}

\bibitem[{Chen et~al.(2020)Chen, Balter, Maguire, and Yarmush}]{Chen2020}
Chen, A., Balter, M., Maguire, T., and Yarmush, M. (2020).
\newblock {Deep learning robotic guidance for autonomous vascular access}.
\newblock \emph{Nature Machine Intelligence} 2, 104--115.
\newblock \doi{10.1038/s42256-020-0148-7}
\bibAnnoteFile{Chen2020}

\bibitem[{Fischler and Bolles(1981)}]{ransac}
Fischler, M.~A. and Bolles, R.~C. (1981).
\newblock Random sample consensus: A paradigm for model fitting with
  applications to image analysis and automated cartography.
\newblock \emph{Commun. ACM} 24, 381–395.
\newblock \doi{10.1145/358669.358692}
\bibAnnoteFile{ransac}

\bibitem[{Gong et~al.(2020)Gong, Jiang, Meng, Ye, Li, Xie et~al.}]{Gong2020}
Gong, Z., Jiang, S., Meng, Q., Ye, Y., Li, P., Xie, F., et~al. (2020).
\newblock {SHUYU Robot: An Automatic Rapid Temperature Screening System}.
\newblock \emph{Chinese Journal of Mechanical Engineering} 33, 38.
\newblock \doi{10.1186/s10033-020-00455-1}
\bibAnnoteFile{Gong2020}

\bibitem[{Henkel et~al.(2020)Henkel, Caic, Blaurock, and Okan}]{Henkel2020}
Henkel, A., Caic, M., Blaurock, M., and Okan, M. (2020).
\newblock {Robotic transformative service research: deploying social robots for
  consumer well-being during COVID-19 and beyond}.
\newblock \emph{Journal of Service Management} \doi{10.1108/JOSM-05-2020-0145}
\bibAnnoteFile{Henkel2020}

\bibitem[{Hollander and Carr(2020)}]{Hollander2020}
Hollander, J.~E. and Carr, B.~G. (2020).
\newblock {Virtually Perfect? Telemedicine for Covid-19}.
\newblock \emph{New England Journal of Medicine} 382, 1679--1681.
\newblock \doi{10.1056/NEJMp2003539}
\bibAnnoteFile{Hollander2020}

\bibitem[{{IGI Testing Consortium}(2020)}]{nature_biotechnology_2020}
{IGI Testing Consortium} (2020).
\newblock {Blueprint for a pop-up SARS-CoV-2 testing lab}.
\newblock \emph{Nature Biotechnology} 38, 791–797.
\newblock \doi{10.1038/s41587-020-0583-3}
\bibAnnoteFile{nature_biotechnology_2020}

\bibitem[{ISO 9283:1998(1998)}]{ISO9283}
ISO 9283:1998 (1998).
\newblock \emph{{Manipulating industrial robots -- Performance criteria and
  related test methods}}.
\newblock Standard, International Organization for Standardization, Geneva, CH
\bibAnnoteFile{ISO9283}

\bibitem[{Li et~al.(2020)Li, Guo, Liu, Wang, Zhou, Yu et~al.}]{Li2001912}
Li, S.-Q., Guo, W.-L., Liu, H., Wang, T., Zhou, Y.-Y., Yu, T., et~al. (2020).
\newblock {Clinical Application of Intelligent Oropharyngeal-swab Robot:
  Implication for COVID-19 Pandemic}.
\newblock \emph{European Respiratory Journal} \doi{10.1183/13993003.01912-2020}
\bibAnnoteFile{Li2001912}

\bibitem[{Quigley et~al.(2009)Quigley, Conley, Gerkey, Faust, Foote, Leibs
  et~al.}]{Quigley2009}
Quigley, M., Conley, K., Gerkey, B., Faust, J., Foote, T.~B., Leibs, J., et~al.
  (2009).
\newblock {ROS}: an open-source {Robot Operating System}.
\newblock In \emph{ICRA Workshop on Open Source Software}
\bibAnnoteFile{Quigley2009}

\bibitem[{{Rublee} et~al.(2011){Rublee}, {Rabaud}, {Konolige}, and
  {Bradski}}]{orb_features}
{Rublee}, E., {Rabaud}, V., {Konolige}, K., and {Bradski}, G. (2011).
\newblock {ORB: An efficient alternative to SIFT or SURF}.
\newblock In \emph{2011 International Conference on Computer Vision}.
  2564--2571
\bibAnnoteFile{orb_features}

\bibitem[{Scassellati and V{\'a}zquez(2020)}]{Scassellatieabc9014}
Scassellati, B. and V{\'a}zquez, M. (2020).
\newblock {The potential of socially assistive robots during infectious disease
  outbreaks}.
\newblock \emph{Science Robotics} 5.
\newblock \doi{10.1126/scirobotics.abc9014}
\bibAnnoteFile{Scassellatieabc9014}

\bibitem[{{United States Centers for Disease Control and
  Prevention}(2020)}]{cdc_2020}
[Dataset] {United States Centers for Disease Control and Prevention} (2020).
\newblock {COVID-19 Death Data and Resources, Daily Updates of Totals by Week
  and State}
\bibAnnoteFile{cdc_2020}

\bibitem[{{World Health Organization}(2020)}]{who_2020}
[Dataset] {World Health Organization} (2020).
\newblock {Coronavirus Disease (COVID-19) Dashboard}
\bibAnnoteFile{who_2020}

\bibitem[{Yang et~al.(2020{\natexlab{a}})Yang, Lv, Zhang, Yang, Deng, You
  et~al.}]{Yang2020}
Yang, G., Lv, H., Zhang, Z., Yang, L., Deng, J., You, S., et~al.
  (2020{\natexlab{a}}).
\newblock {Keep Healthcare Workers Safe: Application of Teleoperated Robot in
  Isolation Ward for COVID-19 Prevention and Control}.
\newblock \emph{Chinese Journal of Mechanical Engineering} 33, 47.
\newblock \doi{10.1186/s10033-020-00464-0}
\bibAnnoteFile{Yang2020}

\bibitem[{Yang et~al.(2020{\natexlab{b}})Yang, J.~Nelson, Murphy, Choset,
  Christensen, H.~Collins et~al.}]{Yangeabb5589}
Yang, G.-Z., J.~Nelson, B., Murphy, R.~R., Choset, H., Christensen, H.,
  H.~Collins, S., et~al. (2020{\natexlab{b}}).
\newblock {Combating COVID-19 {\textemdash} The role of robotics in managing
  public health and infectious diseases}.
\newblock \emph{Science Robotics} 5.
\newblock \doi{10.1126/scirobotics.abb5589}
\bibAnnoteFile{Yangeabb5589}

\bibitem[{Ye et~al.(2020)Ye, Zhou, Shao, Xiong, Hong, Huang et~al.}]{Ye2020}
Ye, R., Zhou, X., Shao, F., Xiong, L., Hong, J., Huang, H., et~al. (2020).
\newblock {Feasibility of a 5G-Based Robot-Assisted Remote Ultrasound System
  for Cardiopulmonary Assessment of Patients With COVID-19}.
\newblock \emph{Chest} \doi{10.1016/j.chest.2020.06.068}
\bibAnnoteFile{Ye2020}

\end{thebibliography}


\section*{Figure captions}

\renewcommand{\arraystretch}{1.5}
\begin{table}[h]
    \centering
    \begin{tabular}{|l|c|c|}
    \hline
    Test Equipment & Manual Operation Time (s) & Teleoperation Time (s)\\
    \hline
        Mock Ventilator & 20 & 74\\
        Mock Ventilator & 18 & 60\\
        Mock Ventilator & 15 & 67\\
        Servo-U Ventilator & 28 & 109\\
    \hline
    \end{tabular}
    
    \caption{Experimental data for manual and teleoperation performance of three routine setting change tasks.}
    \label{tab:times}
\end{table}

\end{document}